\documentclass[10pt,twocolumn,letterpaper]{article}

\usepackage[pagenumbers]{cvpr} 

\usepackage{graphicx}
\usepackage{amsmath}
\usepackage{amssymb}
\usepackage{booktabs}
\usepackage{arydshln}
\usepackage{float}
\usepackage{lipsum}
\usepackage{soul}
\RequirePackage{algorithm}
\RequirePackage[noend]{algorithmic}
\usepackage[pagebackref,breaklinks,colorlinks]{hyperref}

\usepackage[capitalize]{cleveref}
\crefname{section}{Sec.}{Secs.}
\Crefname{section}{Section}{Sections}
\Crefname{table}{Table}{Tables}
\crefname{table}{Tab.}{Tabs.}

\newcommand{\ifor}[1]{{\emph{IFOR}}}
\newcommand{\fort}[1]{{\ifor{}}}

\begin{document}
\author{Ankit Goyal$^{1, 2}$\thanks{Work done while authors were interns at NVIDIA},\ Arsalan Mousavian$^{1}$,\, Chris Paxton$^{1}$,\, Yu-Wei Chao$^{1}$,\, Brian Okorn$^{1, 3*}$ \\
Jia Deng$^{2}$ ,\, Dieter Fox$^{1}$\\
$^{1}$NVIDIA, $^{2}$Princeton University, $^{3}$Carnegie Mellon University\\
}
\title{IFOR: Iterative Flow Minimization for Robotic Object Rearrangement}
\makeatletter
\g@addto@macro\@maketitle{
  \begin{figure}[H]
  \setlength{\linewidth}{\textwidth}
  \setlength{\hsize}{\textwidth}
  \vspace{-10mm}
  \centering
  \resizebox{0.96\textwidth}{!}{\includegraphics[]{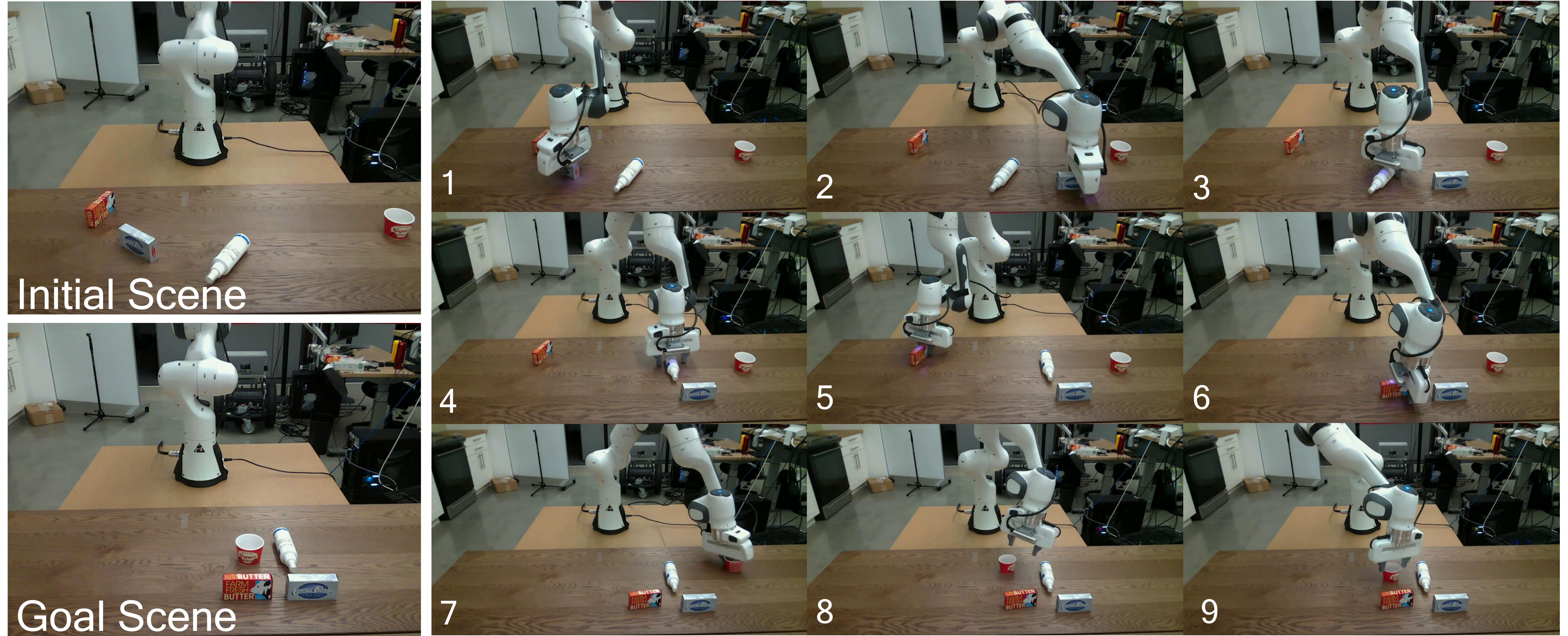}}
  \vspace{-3mm}
  \caption{An example of \fort{} being applied to real data. The initial and goal scenes are shown on the left. Our approach allows the robot to repeatedly identify transformations that will minimize the flow for various objects between the current and goal scenes. It can then repeatedly grasp, move, and place objects, rotating as necessary, in order to achieve the configuration in the goal scene. The system is trained completely on synthetic data and transfers to the real world in zero-shot manner.
  }
  \label{fig:teaser}
  \end{figure}
  \vspace{-2mm}
}
\makeatother

\maketitle
\vspace{-30mm}

\begin{abstract}
\vspace{-2mm}
Accurate object rearrangement from vision is a crucial problem for a wide variety of real-world robotics applications in unstructured environments. We propose \fort{}, Iterative Flow Minimization for Robotic Object Rearrangement, an end-to-end method for the challenging problem of object rearrangement for unknown objects given an RGBD image of the original and final scenes. First, we learn an optical flow model based on RAFT to estimate the relative transformation of the objects purely from synthetic data. This flow is then used in an iterative minimization algorithm to achieve accurate positioning of previously unseen objects. Crucially, we show that our method applies to cluttered scenes, and in the real world, while training only on synthetic data. Videos are available at \url{https://imankgoyal.github.io/ifor.html}.

\end{abstract}

\vspace{-1mm}
\section{Introduction}
\label{sec:intro}
\vspace{-2mm}
Object rearrangement is the capability of an embodied agent to physically re-configure the objects in a scene into a desired goal configuration~\cite{batra:arxiv2020}. It is an essential skill in day-to-day activities like setting a dining table, putting away groceries, and organizing a desk. Endowing robots with this capability is crucial for deploying them to assist people with everyday tasks~\cite{batra:arxiv2020}.

With varying task setups, the desired goal state can be provided in different forms, for instance, a compact state representation~\cite{king:icra2016,weihs:cvpr2021} or natural language descriptions~\cite{shridhar:corl2021,liu:arxiv2021a}.
In this work, we address the rearrangement task where the goal state is specified by an RGB-D image~\cite{labbe:ral2020,qureshi:rss2021}, as shown in Fig.~\ref{fig:teaser}.
This setup lends itself well to many scenarios where the goal state can be snapped once, either in the first place or from a one-time demonstration. For instance, a user can set the dining table once to their preference and take a photo, and a robot assistant can restore the table back to the desired state from any configuration.

Traditionally, object rearrangement problems have been studied in the robotics community, often in the context of Task and Motion Planning (TAMP)~\cite{garrett:arcras2021}. Despite much recent progress~\cite{garrett:icra2020,driess:rss2020,ichter:arxiv2020}, most TAMP approaches still rely on a strict set of assumptions on the perception front. First, the objects and scenes are often assumed to be known a priori, provided with high fidelity 3D models. This makes the approaches difficult to be deployed in unseen environments or environments without models. Second, given the models, the planning front often assumes accurate pose information at the input. This makes explicit object pose estimation~\cite{li:eccv2018,sundermeyer:eccv2018,peng:cvpr2019,wang:cvpr2019,labbe:eccv2020} a necessary part of the pipeline, and the full system susceptible to pose estimation error from real vision systems.

Recent efforts in robotics have attempted to relax these constraints by leveraging the power of deep learning. A recent approach called NeRP, proposed by Qureshi et al.~\cite{qureshi:rss2021}, has allowed for rearranging objects unseen at the training time, by representing the observed objects with learned embeddings. It also removes the need of explicit object pose estimation for planning by leveraging recent progress on learning-based grasp planners~\cite{sundermeyer:icra2021} and collision detectors~\cite{danielczuk:icra2021}.
However, NeRP only allows moving objects with 2D in-plane translations on the table surface and allows no change in their orientation. This prevents its applications in realistic scenarios that require moving objects with more complex transformations such as those shown in Fig.~\ref{fig:teaser}.

We propose a new approach to image-guided robotic object rearrangement with RGB-D input. It achieves, for the first time to the best our of knowledge, the ability to handle unknown objects with translation as well as planar rotations. 

The key to our method is re-formulating object rearrangement as an iterative minimization of optical flow between the current observed image and the goal image. By using optical flow as an intermediate representation, we can capitalize on the cutting edge development in flow estimation models~\cite{teed:eccv2020}. Although these flow models were originally developed for consecutive video frames with small pixel displacements, we show that with proper training, these models can excel at estimating flow with large displacements from arbitrary transformation of objects. Using this estimated flow, together with the depth input and generic object segmentation models~\cite{xiang:cvpr2020}, we can obtain dense 3D correspondences for each object. This provides a general representation that allows us to solve for the desired transformation of objects with simple optimization. Furthermore, with such a general representation, our method trained entirely on synthetic data transfers well to the real world in a zero shot manner. 

To summarize, we introduce~\ifor{}, {\bf I}terative {\bf F}low Minimization for Unseen {\bf O}bject {\bf R}earrangement.~\ifor{} is to our knowledge the first system capable of rearranging unseen objects, given an RGB-D image goal, that handles both translation and rotation. Our approach is trained solely on simulation data and transfers to the real world in a zero-shot manner. Finally, we perform a set of experiments showing our method allows rearrangement of novel objects in cluttered scenes with a real robot.
\section{Related Work}
\begin{figure*}[bt]
\centering
    \resizebox{0.98\textwidth}{!}{\includegraphics{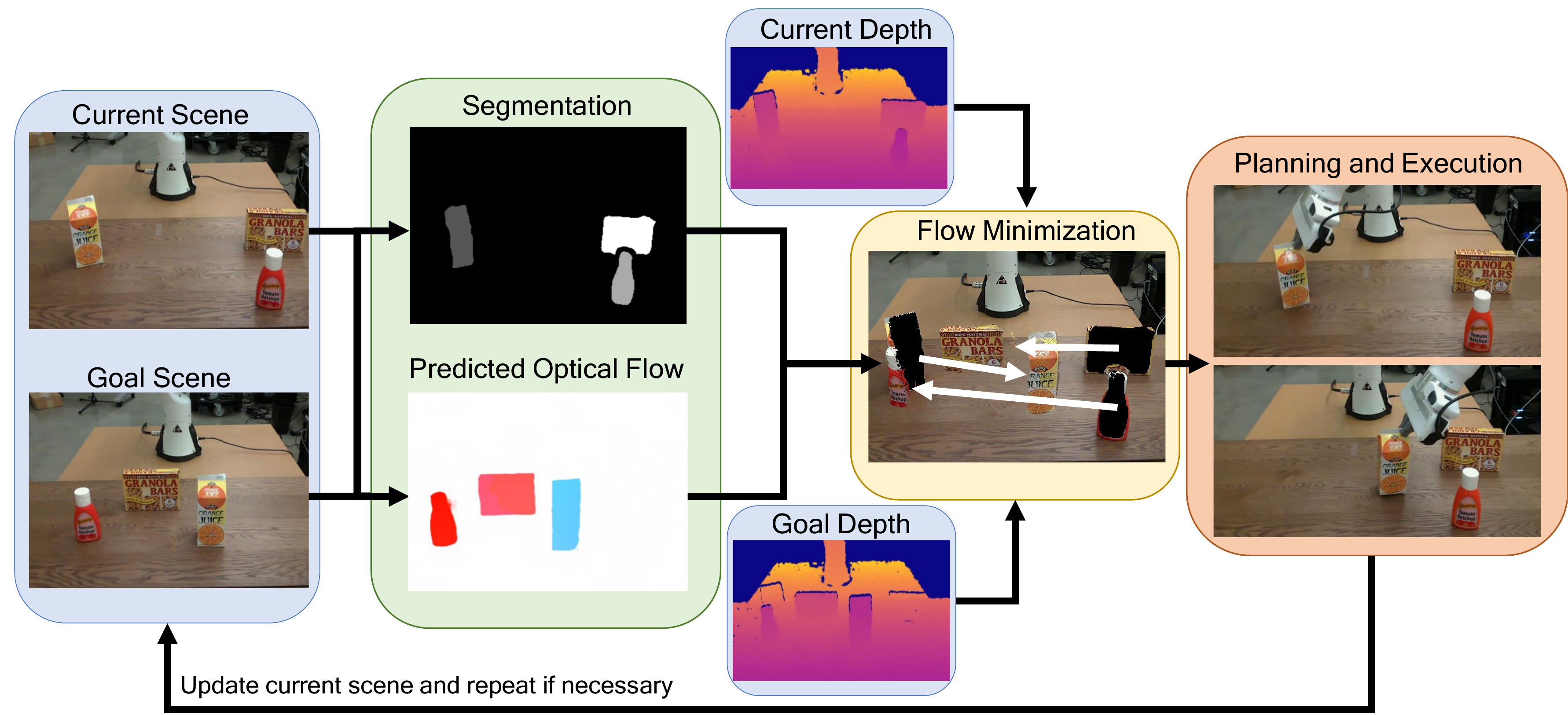}}
    \caption{Overview of the \ifor{} algorithm. \ifor{} takes as input RGB+D images of a current and a goal scene, and uses these to make predicts as to which objects should move and by which transformations, using RAFT to estimate optical flow. This is then sent to a robot planning and execution pipeline which is capable of grasping of unknown objects and motion planning in scenes with unknown geometry.}
    \label{fig:system}
\end{figure*}

\label{sec:background}
\paragraph{Robotic Object Manipulation.}~Our work falls in the broad area of robotic manipulation. Conventional manipulation systems often take a modular approach, decomposing the full system into perception, planning, and actuation components. The perception module is charged with estimating the state of the environment, e.g., detecting and segmenting objects~\cite{he:iccv2017,danielczuk:icra2019,carion:eccv2020,xiang:corl2020,liu:iccv2021} and estimating their 6D poses~\cite{li:eccv2018,sundermeyer:eccv2018,peng:cvpr2019,wang:cvpr2019,labbe:eccv2020}. With the perception output, the planning module then searches for a sequence of actions to accomplish the manipulation task. In robotics, this is conventionally formalized and studied in the problem of Task and Motion Planning (TAMP)~\cite{garrett:icra2020,driess:rss2020,ichter:arxiv2020,garrett:arcras2021}. However, setting up a real-world TAMP system often requires substantial task-specific knowledge and accurate 3D models of the environment, significantly limiting the environments to which the system can generalize. 
To address this challenge, recent work has adopted deep learning-based approaches for robotic manipulation, for instance, on grasp planning~\cite{mahler:rss2017,morrison:rss2018,mousavian:iccv2019,sundermeyer:icra2021,wang:iccv2021}, motion planning~\cite{simeonov:corl2020,danielczuk:icra2021}, and reasoning about spatial relations~\cite{goyal:neurips2020,paxton:corl2021,liu:arxiv2021a}.

Our work is concerned with rearranging objects, an area that has a long history in robotics~\cite{king:icra2016,labbe:ral2020,seita:icra2021,qureshi:rss2021,kapelyukh:corl2021} but has recently gained traction in the vision and learning communities~\cite{batra:arxiv2020,goyal:icml2020,weihs:cvpr2021,liu:arxiv2021b} thanks to the advances in simulation platforms. The works most relevant to ours are that of Labb{\'{e}} et al.~\cite{labbe:ral2020} and NeRP~\cite{qureshi:rss2021}, which also address the rearrangement task with the goal state specified by an image. A key step to address this problem is to establish the correspondence of objects between the current and goal image and solve for their desired transformations. However, both~\cite{labbe:ral2020} and~\cite{qureshi:rss2021} only consider
2D planar translation with no orientation change.
This is arguably because their feature descriptors for objects are generic and not rotation sensitive. In contrast, we propose to use optical flow as the low-level feature descriptors, which can be naturally used to infer the full 6D transformations. In parallel to our work, recent efforts have also addressed rearrangement particularly learned from human demonstrations~\cite{florence:corl2018,zeng:corl2020} and also with different goal specifications such as language~\cite{liu:arxiv2021a,shridhar:corl2021}.

\vspace{-3mm}
\paragraph{Optical Flow and Feature Correspondence.}~Optical flow is a long-standing vision problem that addresses the estimation of pixel motions between two video frames. Alike other sub-fields in computer vision, the take-off of deep learning has replaced the traditional pipelines for optical flow with learning-based end-to-end architectures~\cite{ilg:cvpr2017,sun:cvpr2018,hur:cvpr2019,teed:eccv2020,aleotti:cvpr2021,luo:cvpr2021,stone:cvpr2021,sun:cvpr2021,jiang:iccv2021,zhang:iccv2021}. In tackling object rearrangement, our work proposes to use the predicted optical flow between the current and goal image to establish the desired object transformations. However, rather than estimating flow from consecutive video frames, we demonstrate that state-of-the-art models can excel at predicting flow with large displacements from arbitrary goal images in object rearrangement.

In addition to optical flow, our work can potentially also leverage the body of work on scene flow, which directly predicts the motion in 3D rather than in the 2D image space. Recent work has studied various setups for scene flow including predicting from monocular frames~\cite{hur:cvpr2020}, stereo images~\cite{behl:iccv2017,ma:cvpr2019,jiang:iccv2019}, RGB-D pairs~\cite{lv:eccv2018,teed:cvpr2021,hur:cvpr2021}, to 3D point clouds~\cite{gu:cvpr2019,liu:cvpr2019,mittal:cvpr2020,gojcic:cvpr2021}. Finally, our flow prediction task is closely related to the problem of establishing feature correspondences~\cite{yi:cvpr2018,dusmanu:cvpr2019,sarlin:cvpr2020,zhou:cvpr2021} in 3D reconstruction (e.g., SfM) and visual localization (e.g., SLAM).

\vspace{-3mm}
\paragraph{Embodied AI.}~Our work fits well with recent trends in embodied AI. Initially, this work largely centered around the family of tasks pertained to navigation~\cite{das:cvpr2018,xia:cvpr2018,savva:iccv2019,deitke:cvpr2020,shridhar:cvpr2020}, but has gradually grown to encompass tasks pertained to physical manipulation~\cite{james:ral2020,xiang:cvpr2020,ehsani:cvpr2021}. Moreover, a recent notable article by Batra et al.~\cite{batra:arxiv2020} has recognized the rearrangement problem as a ``canonical task'' for evaluating embodied AI. Our work drives progress in this exact frontier. And unlike most prior embodied AI works, which run evaluation only in simulation, we also evaluate the performance of our method on a real-world robotic platform.
\section{Method}
\ifor{} takes as input the RGB-D images of the current and the goal scene, and iteratively generates a pick-and-place action for one object at a time. At each iteration, the RGB-D image of current and goal scene is passed through two components: (1) perception and (2) planning (Fig.~\ref{fig:system}).

The perception component is responsible for estimating the relative transformation of all objects between the current and goal scene. Given the estimated transforms, the planning component selects an object to be moved along with the required transformation, by taking into account collision and kinematic feasibility. Finally, after executing the planned pick-and-place action, the system will take a new observation of the scene and repeat the process.

\begin{figure*}[t]
    \centering
    \resizebox{\textwidth}{!}{\includegraphics{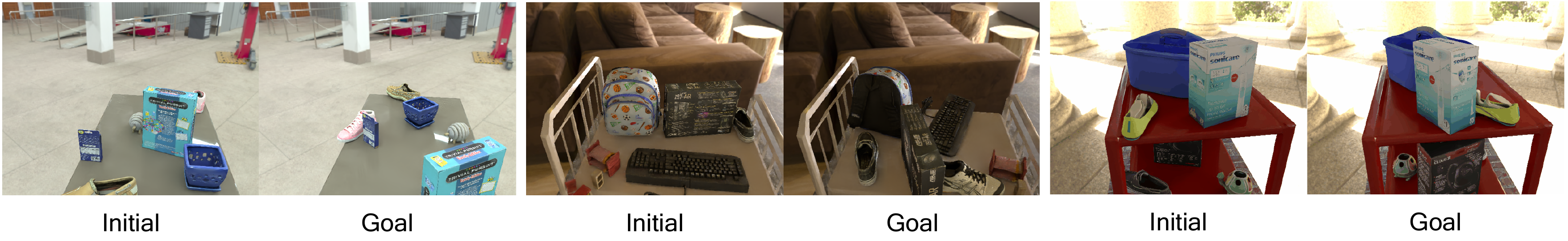}}
    \caption{Examples from the synthetic dataset used for training RAFT. Images come in pairs of initial and final images, each containing assorted objects in clutter. Between the two images, there is a large, randomly-sampled transformations. Retraining RAFT on these large discontinuities is essential to our approach.}
    \label{fig:syn_samples}
    \vskip -0.25cm
\end{figure*}

\subsection{Perception}
The task of the perception component is to segment out objects in the current and goal images, establish their correspondences, and predict a 6-DoF transformation for each object from its current pose to its goal pose. We achieve this with a pipeline that combines optical flow estimation, unseen object segmentation, and a RANSAC-based transformation optimization.

\vspace{-3mm}
\paragraph{Optical Flow For Rearrangement.} The first step is to estimate the optical flow between the current and the goal image. This provides us with a pixel-level correspondence between them. In conventional settings, optical flow is generally estimated between temporally close images in a video. The displacement of the flow is often small as temporally close images do not differ much from one another. In fact, this small displacement prior is crucial for classical methods like Lucas-Kanade method~\cite{lucas1981iterative}. This assumption however does not hold in rearrangement, as the objects could be moved a large distance and rotated by large angles from the initial scenes. This makes classical approaches of optical flow estimation ill-suited for rearrangement.

Contrary to classical methods, deep learning based optical flow methods like Recurrent All Pairs Field Transforms (RAFT)~\cite{teed:eccv2020} learn to make predictions from data. But, they too were developed for and trained on estimating flow in videos and hence pre-trained RAFT model does not work well on rearrangement scenes. 
However, RAFT's underlying architecture does not make any assumption about the flow displacement being small, as it compares every pixel in one image to every other pixel in the other image. Consequently, given suitable data, RAFT can be trained for object rearrangement with large translations and rotations. 

\vspace{-3mm}
\paragraph{RAFT.} Recurrent All Pairs Field Transforms (RAFT) estimates optical flow by constructing a 4D correlation volume of which each pixel in one image is compared to every pixel in the other image~\cite{teed:eccv2020}. It then updates the flow estimates using a recurrent unit, starting with zero optical flow at all locations. In each iteration, the recurrent unit does a lookup around the current estimate of flow to decide how to update the flow estimates. 

During training, RAFT applies supervision over all these intermediate flow estimates made by the recurrent unit. Suppose, $\{\mathbf{f}_1, ..., \mathbf{f}_N \}$ are the $N$ intermediate flow estimates and $\mathbf{f}_{gt}$ is the ground-truth flow, then the loss ($\mathcal{L}$) is defined as the weighted sum of $l_1$ distance between estimated and ground-truth flow. Specifically, 
\begin{equation}
    \mathcal{L} = \sum_{i=1}^{N} \gamma^{N-i} ||\mathbf{f}_{gt} - \mathbf{f}_i||_1,
\end{equation}
where $\gamma$ is a discount factor of value 0.8. For more details, please refer to the work by Teed an Deng~\cite{teed:eccv2020}.

\vspace{-3mm}
\paragraph{Synthetic Data.}
To train RAFT for object rearrangement, we create a visually realistic dataset of synthetic scenes. In these scenes, objects are placed on supports like tables or beds. We sample the supports from ShapeNet~\cite{chang:arxiv2015} and the objects from the Google Scanned Dataset~\cite{gso:2021}. We use the NViSII renderer~\cite{morrical:2020} to render scenes with realistic lighting via ray tracing, and
diverse view points by randomizing camera poses.
We also randomize the lighting of the scenes, the texture of the supports and the background image. In the end, we created a dataset of $\sim$54K training samples and 1000 test samples. Some samples of our synthetic data are shown in Fig.~\ref{fig:syn_samples}. 

\vspace{-3mm}
\paragraph{Unseen Object Segmentation.}
Optical flow alone is not sufficient to estimate the transformation of the object as they lack the grouping or ``objectness" information. Moreover, unlike most prior segmentation methods that only segment object classes they were trained on, we need a zero-shot segmentation method to deal with unseen objects at test time. We use pre-trained UCN~\cite{xiang:corl2020}, which is trained to segment unknown objects from RGB-D image of the scene. It learns per-pixel embedding such that pixels belonging to the same object instance have similar embeddings but different from other object instances in the scene. At the test time, objects are segmented using mean-shift clustering.

\vspace{-3mm}
\paragraph{Relative Object Transform from Flow.}
The next step is to estimate the relative transformation of the objects between the two frames. The relative transformation of each object is computed by first unprojecting each pixel to 3D from the depth map of the scene and camera intrinsics. The 3D correspondence between current image and goal image is computed from predicted flow and unprojected points.

\begin{figure*}[bth!]
    \resizebox{\textwidth}{!}{\includegraphics{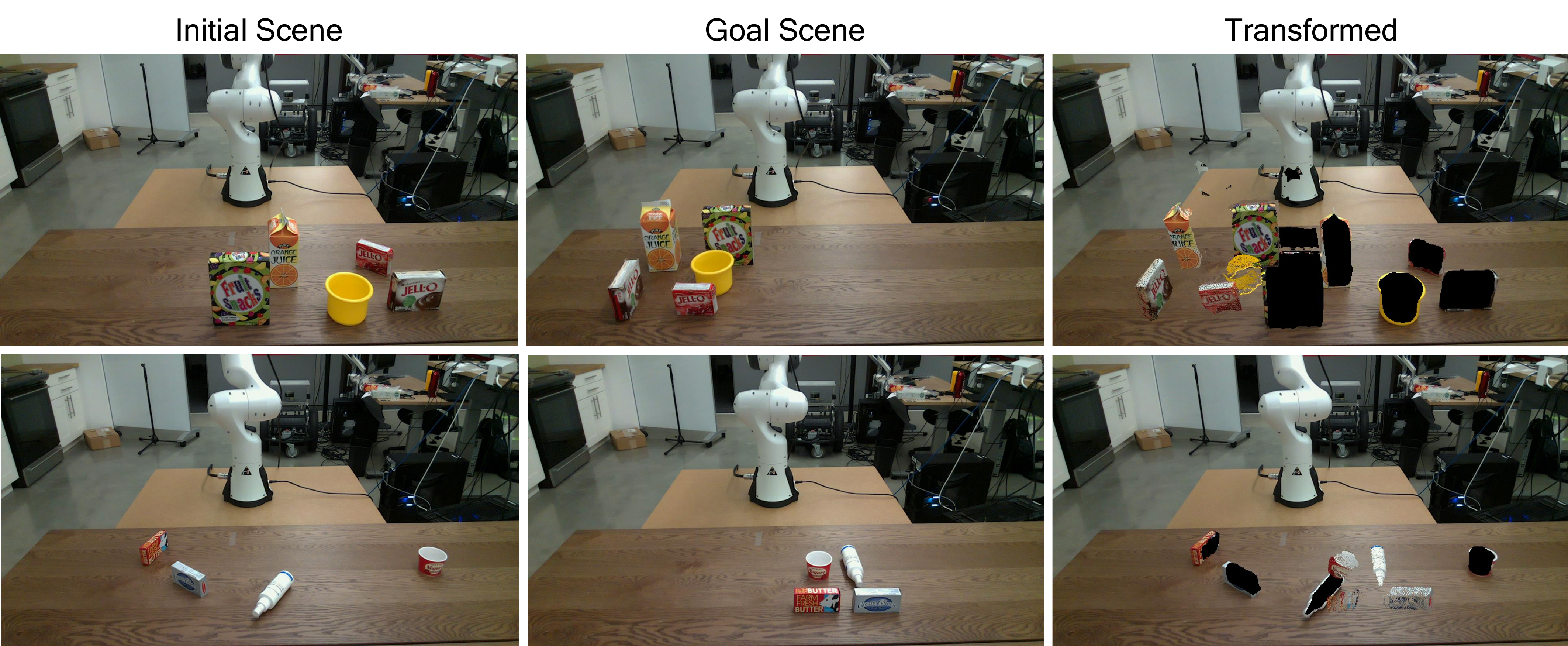}}
    \caption{Examples of different real-world scenes warped after object points have been transformed according to poses estimated by \ifor{}. Poses are estimated according to predicted optical flow features.}
    \label{fig:flow-example}
    \vskip -0.25cm
\end{figure*}
\begin{algorithm}[t]
    \begin{algorithmic}[1]
        \STATE {\bfseries Input:} Objects [$O_1, \dots, O_n$], Estimated transformation [$T_1, \dots, T_n$]  
        \FOR{$i=1$ {\bfseries to} $n$}
        \STATE Calculate score $S_i$ for object $O_i$
        \ENDFOR
        \STATE Sort objects [$O_1, \dots, O_n$] and transformations [$T_1, \dots, T_n$] based on score [$S_1, \dots, S_n$] 
        \STATE Let sorted object be [$\bar{O}_1, \dots, \bar{O}_n$] and sorted transformations be [$\bar{T}_1, \dots, \bar{T}_n$]
        \STATE ObjectMoved = False
        \FOR{$i=1$ {\bfseries to} $n$}
        \STATE Collision = if $\bar{O}_i$ collides when applied transformation $\bar{T}_i$
        \IF{Collision is False}
            \STATE Move object $\bar{O}_i$ with relative transform of $\bar{T}_i$
            \STATE ObjectMoved = True
            \STATE break loop
      \ENDIF
      \ENDFOR
      \IF{ObjectMoved is False}
            \FOR{$i=1$ {\bfseries to} $n$}
            \STATE FreeSpaceFound = Find free space $\bar{F}_i$ to place $\bar{O}_i$
            \IF{FreeSpaceFound is True}
                \STATE Move $\bar{O}_i$ to $\bar{F}_i$
                \STATE break loop
            \ENDIF
            \ENDFOR
      \ENDIF
    \end{algorithmic}
    \caption{Pseudo Code for Planning Module in IFOR}
    \label{alg:planning}
\end{algorithm}

We solve for a rigid body transformation for each object by minimizing the error in position after applying the transformation. Let $P_c\in\mathbb{R}^{3\times n}$ be the 3D points of an object in the current scene and $P_g\in\mathbb{R}^{3\times n}$ be the corresponding points in the goal scene. We then estimate a rigid-body rotation $R$ and translation $T$ by solving the following optimization problem using SVD decomposition:
\begin{equation}
    \mathop{\arg \min}\limits_{R, T} ||(R\cdot P_c + T) - P_g||_2.
\end{equation}

In practice, we observed that the matched correspondences from flow can contain many outliers. This is especially prominent when the object undergoes extreme transformations (e.g., large rotations), resulting in only a small portion of the object surface that is visible in both images. To handle the outliers reliably, we adopt RANSAC~\cite{fischler:cacm1981} in solving the relative pose. In Table~\ref{tab:opt_vs_learning}, we show that RANSAC is effective in removing outliers and estimating accurate transformations. In Figure~\ref{fig:flow-example}, we show qualitative examples of the transformations estimated by RANSAC using our trained RAFT model.

\begin{figure*}[bt]
    \resizebox{0.96\textwidth}{!}{\includegraphics{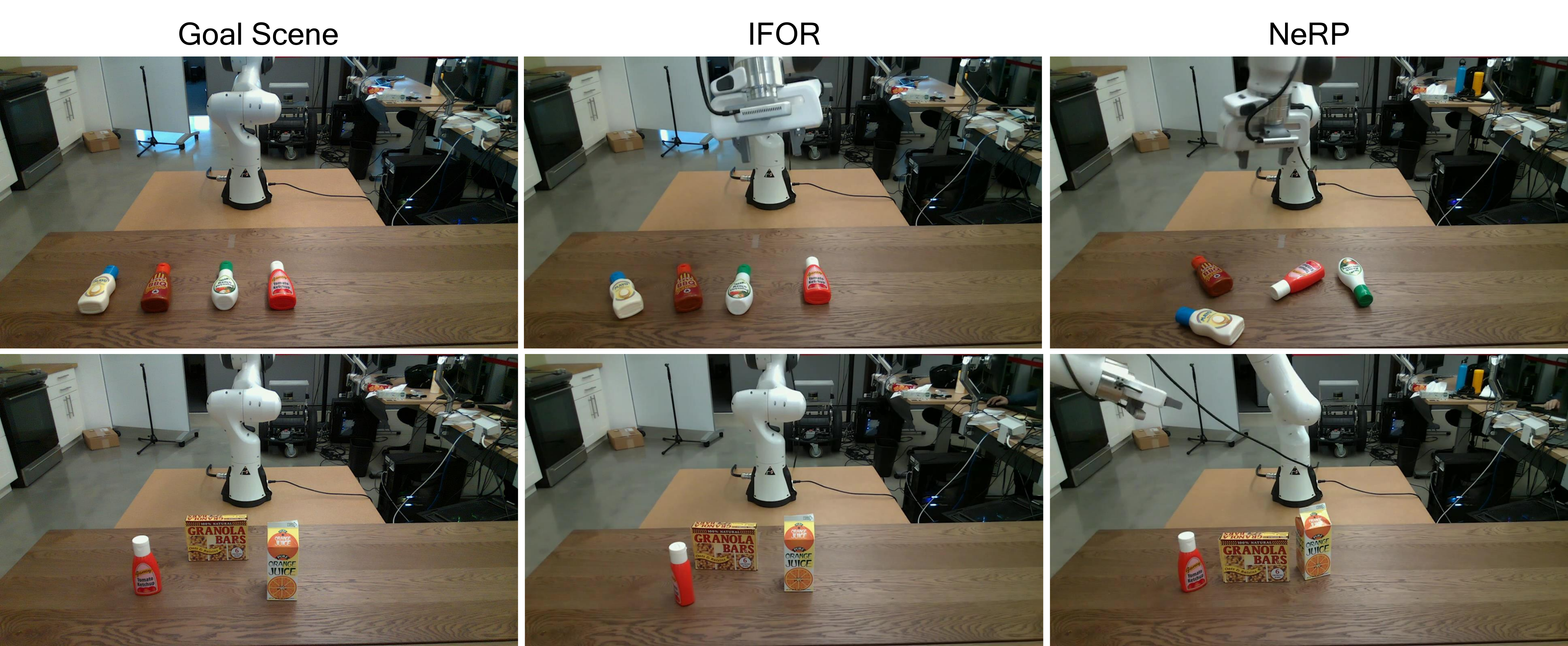}}
    \caption{Two examples comparing the qualitative performance of NeRP~\cite{qureshi:rss2021} to \ifor{}. In both of these examples, \ifor{} both more closely matches the goal image, and matches the orientation much more precisely than NeRP does.}
    \vspace{-2mm}
    \label{fig:ifor_vs_nerp}
\end{figure*}

\subsection{Planning and Execution}
\vspace{-2mm}
Given the list of desired object transformations, the planner produces a pick-and-place action to execute, while taking into account various kinematic and geometric constraints. Our planning algorithm iterates over the list of desired object transformation from perception module, and finds out which objects can be moved directly using the predicted transform. The planner classifies each relative transform as feasible if the object at the predicted transform is not colliding with any other objects in the scene. We used pre-trained SceneCollisionNet~\cite{danielczuk:icra2021} to check the collision of the object at the predicted transformation. The feasible objects are then ranked based on score $S = |r| + \lambda |t|$, where r is the relative rotation transformation in radians, t is the relative translation in cm and $\lambda = 0.2$ in our experiments. The planning algorithm greedily picks objects with larger relative transformations.

\vspace{-1mm}
If the policy is not able to find a feasible movement for any object, it will try to move one of the objects to a random collision-free location. Provided that the estimated transformation of the objects are correct, the proposed planning and execution policy is guaranteed to converge and successfully rearrange the objects, i.e., in the worse case, it will move all but one object to collision-free locations, followed by moving each of them to the goal location one by one. 

\vspace{-1mm}
The system  terminates  when  the  estimated  change  in  rotation and translation for all objects is smaller than than a fixed threshold.  Specifically, we use a threshold of $10^\circ$ for rotation and $5 \ cm$ for translation. Our experiments show that this heuristic is quite effective in handling challenging object rearrangement scenarios in the real world (discussed in Sec.~\ref{sec:real}). The pseudo-code for the planning algorithm is outlined in Algo.~\ref{alg:planning}.
\begin{figure}[bt]
    \resizebox{0.49\textwidth}{!}{\includegraphics{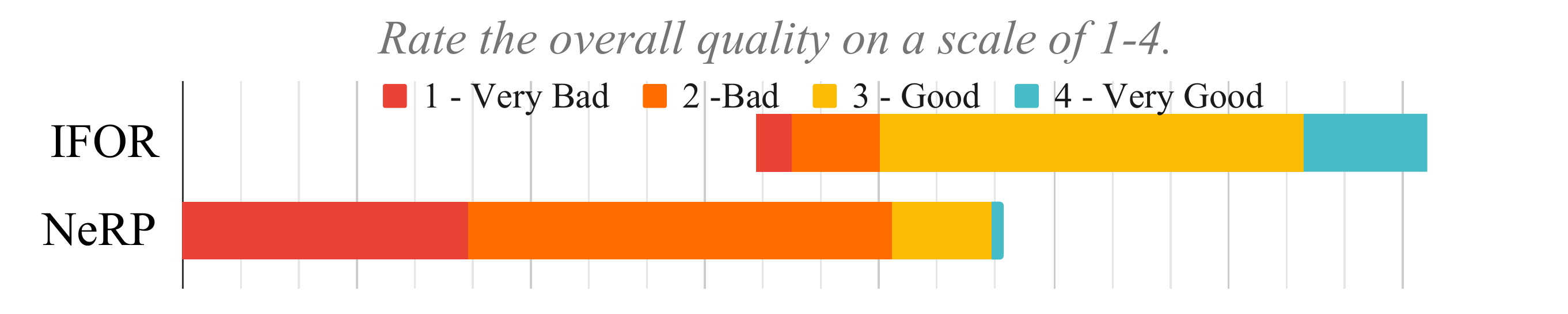}}
    \resizebox{0.49\textwidth}{!}{\includegraphics{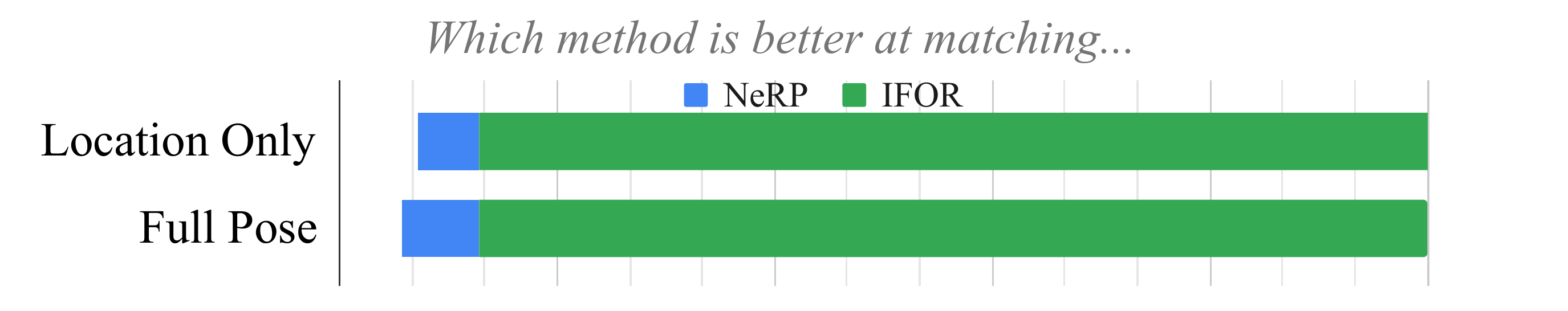}}
    \caption{User scores for \ifor{} vs. NeRP~\cite{qureshi:rss2021}. When asked to rate performance of the two methods on a scale of 1-4, users preferred \ifor{} by a wide margin. Users chose \ifor{} over NeRP in almost all situations, when looking at either position only (94\%) or full pose (position and orientation, 92\%).}
    \label{fig:user-scores}
    \vskip -0.5cm
\end{figure}

\begin{figure*}[t]
    \centering
    \resizebox{0.96\textwidth}{!}{\includegraphics{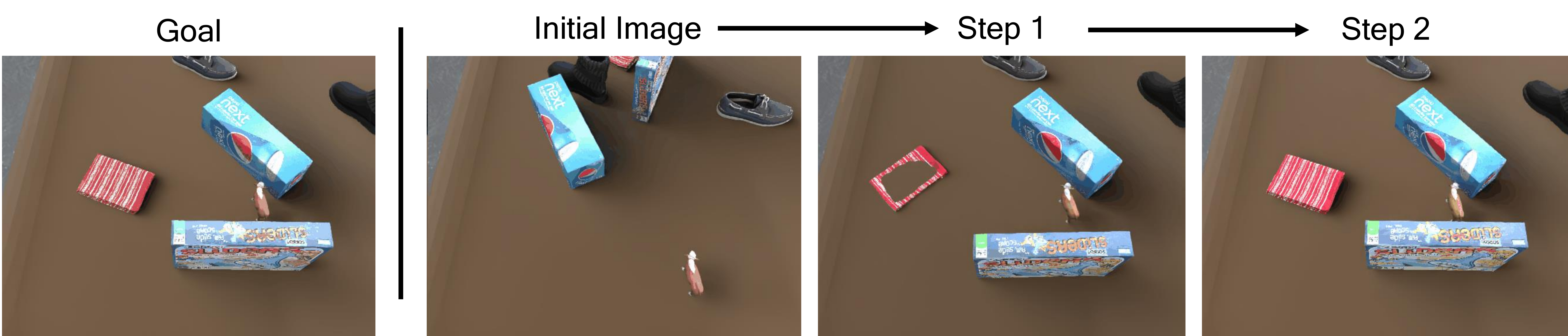}}
    \caption{An example showing the sequence of transformations executed by our model on the synthetic data. Objects are teleported into the next position as predicted by IFOR without collision checking. This figure only shows the first two steps, but minor refinements are still possible.}
    \label{fig:teleport}
    \vskip -0.25cm
\end{figure*}

\begin{table*}[bht]
    \centering
    \begin{tabular}{ccccc}
    \toprule
    & \multicolumn{2}{c}{Rot. $\in [-60^{\circ}, 60^{\circ}]$} & \multicolumn{2}{c}{Rot. $ \in [-180^{\circ}, 180^{\circ}]$} \\
    \cmidrule(lr){2-3}
    \cmidrule(lr){4-5}
    & Median Rot. Err.  & Median Pos. Err. & Median Rot. Err. & Median Pos. Err. \\
    Method & (in $^\circ$) & (in cm) & (in $^\circ$) & (in cm)\\
    \midrule
    Learning Baseline & 22.8 & 8.1 & 33.8 & 8.2 \\
    Pretrained RAFT + RANSAC & 20.6 & 46.5 & 67.8 & 43.1  \\
    Rearrangement RAFT + RANSAC   & {\bf 3.6} & {\bf 1.2} & {\bf 13.7} & {\bf 2.7} \\
    \bottomrule
    \end{tabular}
    \caption{Performance of various approaches in estimating pose transformation from flow in one-step on simulated data. We find that RANSAC outperforms the learning based method for finding the relative transform. In addition, pre-trained RAFT model performs poorly on the rearrangement scenes where the object displacements are larger than  video sequences.}
    \label{tab:opt_vs_learning}
    \vskip -0.25cm
\end{table*}

\section{Experiments}
We present results in two settings. First, we perform an integrated system comparison in the real world, with a physical robot picking and placing objects using either \ifor{} or the previous state of the art~\cite{qureshi:rss2021}. Second, we perform large scale ablation studies in simulation to evaluate the effect of each design choice in \ifor{}. 

\subsection{Real World Experiments}
\label{sec:real}
We used a Franka Panda robot to conduct the real world experiments. The world is perceived from an external RealSense L515 camera, and a wrist-mounted RealSense D415 camera. The external camera is used for planning with \ifor{}, collision avoidance, and controlling the robot. The wrist mounted camera is only used for grasping to improve the robustness of the system.

\ifor{} generates a list of objects and transforms representing the placement location. These poses are passed to the pick-and-place system which takes the ordered list of actions from the planner. It selects the first object that can be grasped and places it at the desired location. Grasps for each object are computed by Contact-GraspNet~\cite{sundermeyer:icra2021} and the robot motions are generated using a model predictive control pipeline and SceneCollisionNet. Refer to \cite{danielczuk:icra2021} for details of the pick and place system. None of the components are trained on any real objects nor the objects we tested on.

We evaluated on 6 scenes, where each scene has between 2 to 5 objects in the initial configuration and a distinct goal configuration. Although it is not possible to replicate the exact initial conditions in the real world, we tried to duplicate the initial configuration visually, and used the same RGB-D goal image for testing both methods.
In order to quantitatively evaluate the performance of the methods in the real world, we conducted a user study with 10 users, where we asked users to select the method that performed better: \ifor{} or NeRP~\cite{qureshi:rss2021}. We also asked users to rate \ifor{} and NeRP on a scale of 1-4, where 1 is ``very bad`` and 4 is ``very good``. 

All the components of the pick-and-place system were the same for both \ifor{} and NeRP, except the estimation of the objects' final pose.
Fig.~\ref{fig:ifor_vs_nerp} shows a qualitative comparison of \ifor{} and NeRP given similar target images. The video of the experiments are included in the Supplementary Materials. Since NeRP does not handle changing the orientation, we asked users to rank the two methods only based on translation as well as considering both rotation and translation. Fig.~\ref{fig:user-scores} shows that users significantly prefer \ifor{} over NeRP both on the translation only setting and full pose setting. NeRP failures were often due to an incorrect correspondence between the current image and target image. In addition, the learned placement generator does not generalize well to arbitrary rearrangements. \ifor{}, on the other hand, can find corresponding objects and their relative pose transform reliably from the predicted optical flow. 

\begin{figure*}[bht!]
\centering
\resizebox{0.40\textwidth}{!}{\includegraphics{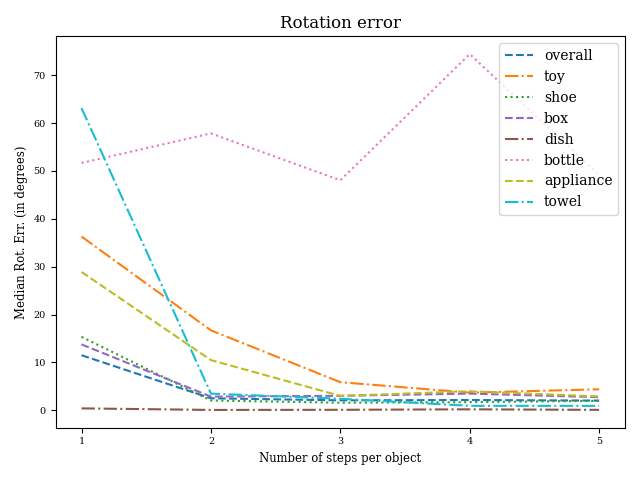}}
\resizebox{0.40\textwidth}{!}{\includegraphics{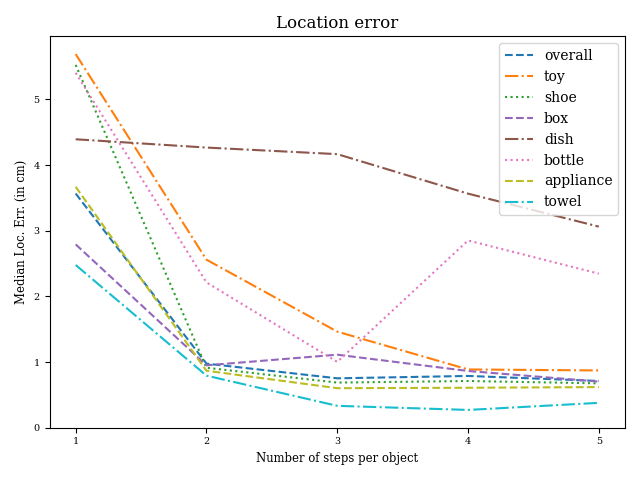}}
\caption{Per-class results over time, showing both position and rotation error. Objects from the Google Scanned Objects dataset~\cite{gso:2021} were divided into different groupings containing at least 10 distinct models. This class breakdown shows that the system has the most trouble with object classes that have a lot of physical symmetry; the ``bottle'' category includes objects where the only orientation distinction is often a label.}
\label{fig:by_class}
\vskip -0.25cm
\end{figure*}

\begin{table*}[tbh!]
  \centering
  \setlength{\tabcolsep}{5pt}
  \begin{tabular}{ccccccccc}
    \toprule
    GT Coll. & Learned     & Planar      & & Median           & Median       &  $|\Delta t| < 2cm$     & $|\Delta t| < 5cm$      & $|\Delta t| < 10cm$ \\
    Check    & Coll. Check & Rot. & Planning & $|\Delta \theta|$ (in $^\circ$) & $|\Delta t|$ (in cm) & $|\Delta \theta| < 5^\circ$ & $|\Delta \theta| < 10^\circ$ & $|\Delta \theta| < 15^\circ$\\
    \midrule
     &    & \checkmark &  & 1.3 & 0.58 & 72.8\% & 78.4\% & 79.5\% \\
     \midrule
    
    \checkmark  &    & \checkmark & \checkmark & 1.4 & 0.64 & 70.0\% & 81.8\% & 84.0\% \\
    \checkmark  &    &  & \checkmark & 3.5 & 1.05 & 59.7\% & 74.2\% & 76.9\%  \\
     &  \checkmark   & \checkmark & \checkmark & 1.6 & 1.09 & 49.2\% & 64.7\% & 68.1\%   \\
     &  \checkmark   &  & \checkmark & 4.5 & 2.04 & 37.0\% & 53.5\% & 57.9\% \\
    \bottomrule
  \end{tabular}
\caption{Ablation results for IFOR in simulation. First row corresponds to the teleportation baseline discussed in Sec.4.2, and serves as an no-planning upper-bound. Our planning policy achieves close to this no-planning upper-bound. Further, we show that IFOR benefits from better collision detection and assumptions about planar rotations.}
\label{tab:simulation}
\vskip -0.25cm
\end{table*}

\subsection{Ablation Studies}
We conducted our ablation studies on a synthetic dataset which consisted of 200 tabletop scenes with randomized lighting, texture and background images. Each scene contained between 1 and 9 objects. Given a random target scene, the current scene was generated by applying a random planar rotation and translation to each object. We first define the evaluation metrics below and then provide the analysis on different ablation studies.

\vspace{-3mm}
\paragraph{Metrics.}
We report the median translation and rotation error averaged over all the objects. We also report the percentage of objects that are within a threshold of rotation and position error for different thresholds. In addition, for Table~\ref{tab:simulation}, we divide the scenes by three task difficulty levels: ``easy`` for position error less than 2 cm and rotation error less than $5^\circ$; ``medium`` for position error less than 5 cm and rotation error less than $10^\circ$; and ``hard`` for position error less than 10 cm and rotation error less than $15^\circ$. 

\vspace{-3mm}
\paragraph{Learning-based versus RANSAC for relative transform prediction.}
We explored learning based solution to predict transformation from flow instead of optimizing it with RANSAC. For estimating location, we predict the heat-map of the centroid of the object in the goal scene given the flow and depth image of the scene. For estimating rotation, we use a cropped image of the flow around the object and regress the change in rotation. Table~\ref{tab:opt_vs_learning} compares the accuracy of the learning-based baseline with our RANSAC-based optimization. Our RANSAC-based optimization with re-trained RAFT (rearrangement RAFT) achieves significantly lower errors.

\vspace{-3mm}
\paragraph{Pre-trained RAFT versus Rearrangement RAFT.} We compared the accuracy of single step transformation prediction between pre-trained RAFT which is trained on video sequences vs rearrangement RAFT where it is trained on our synthetic data with large translation and orientation changes. Table~\ref{tab:opt_vs_learning} shows that training RAFT on rearrangment scenes is cruicial for achieving high accuracy for predicting relative transforms of the objects.

\vspace{-3mm}
\paragraph{Teleporting objects versus planning one action at the time.} In order to provide an approximate upper-bound performance of \ifor{}, we implemented a baseline where the policy computes transformations for all the objects and move all of them at once and observe the scene again. Fig.~\ref{fig:teleport} shows the execution of teleport policy in simulation. This policy is clearly not practical in any real robotic setting but removes planning constraints and achieves significantly better results as shown in first row of Table~\ref{tab:simulation}. In addition, we can see the the heuristic planning does not lose the performance compared to the teleport policy which shows its effectiveness.

\vspace{-3mm}
\paragraph{Ground-truth Collision Checker versus Learned Collision Checker~\cite{danielczuk:icra2021}.} An important component of \ifor{} is the collision checker which indicates whether a predicted rotation and translation should be accepted by the policy or not. Table~\ref{tab:simulation} shows that the placement accuracy drops due to imperfections of the learned collision checker. \ifor{} benefits from improvements in collision checking for arbitrary scenes.

\vspace{-3mm}
\paragraph{Planar Rotation versus SO(3) Rotation.} Given the correspondences in 3D, we optimize for relative $SE(3)$ transform using RANSAC. However, we can use the knowledge that objects rotations are planar and only keep the rotation component around z-axis. Table~\ref{tab:simulation} shows that adding planar rotation assumption improves the accuracy significantly. 

\vspace{-3mm}
\paragraph{Object-wise Performance Analysis.}
In order to get more insight into the performance of the system, we divided the objects of \cite{gso:2021} into 7 classes that are shown in Fig.~\ref{fig:by_class}. The general trend among the categories is that the accuracy improves with the increase in the number of steps which stems from having more overlapping views of the object. The category that \ifor{} struggles with most is \textit{bottle}. Bottles are challenging because the method needs to match not only the geometry of object but also needs to match the texture details on the surface of the bottles, which can be quite challenging for smaller objects where there is not enough resolution on those objects. We can see a similar issue with the \textit{dish} category, which includes pots, pans, and mugs.
Unlike with bottles, these objects tend to have little to no texture and are mostly symmetric. 

\section{Discussion}

We proposed \ifor{}, an iterative optical flow based system for object rearrangement. To the best of our knowledge, \ifor{} is the first method that can solve the image-based object rearrangement task, with both translation and rotation, for unseen objects.
We conducted experiments with a robot in the real world to show its effectiveness and generalization to real perception data. Our results show a significant performance boost over previous work. In addition, we conducted a comprehensive analysis on  simulation data to evaluate the effect of each component of our approach. 

There are multiple promising future directions for this work. The first is to extend \ifor{} to $SE(3)$ poses. Even though there is no inherent assumption about planar rotation in our method, we only evaluated the object rearrangement problem with planar rotations because the models are only trained on data with planar rotations.

Another interesting direction would be to monitor the scene during robot execution to account for any in-gripper object motions and update the placement pose accordingly.

{\small
\bibliographystyle{ieee_fullname}
\bibliography{egbib}
}

\end{document}